\documentclass[11pt,onecolumn]{article}
\usepackage{a4wide}
\usepackage[]{amssymb}
\usepackage[]{amsmath}
\usepackage[utf8]{inputenc}
\usepackage{graphicx,color}

\title{Modeling the growth of fingerprints\\improves matching for adolescents}
\author
{Carsten Gottschlich\thanks{C.~Gottschlich acknowledges support by DFG RTS 1023, T.~Hotz by DFG CRC 803.} \thanks{C.~Gottschlich, T.~Hotz and A.~Munk are with the Institute for Mathematical Stochastics, University of G\"ottingen, Goldschmidtstrasse 7, 37077 G\"ottingen, Germany.
E-mails:  \{gottschlich,hotz,munk\}@math.uni-goettingen.de} , Thomas Hotz$^{*\dag}$, Robert Lorenz\thanks{R.~Lorenz, S.~Bernhardt and M.~Hantschel are with the Bundeskriminalamt, Wiesbaden, Germany.} , \\Stefanie Bernhardt$^{\ddag}$, Michael Hantschel$^{\ddag}$, and Axel Munk$^{\dag}$
}

\begin{document}

\maketitle

\begin{abstract}

We study the effect of growth on the fingerprints of adolescents, based on which we suggest a simple method to adjust for growth when trying to recover a juvenile's fingerprint in a database years later.
Based on longitudinal data sets in juveniles' criminal records, we show that growth essentially leads to an isotropic rescaling, so that we can
use the strong correlation between growth in stature and limbs to model the growth of fingerprints proportional to stature growth as documented in growth charts.
The proposed rescaling leads to a 72\% reduction of the distances 
between corresponding minutiae for the data set analyzed.
These findings were corroborated by several verification tests.
In an identification test 
on a database containing 3.25 million right index fingers
at the Federal Criminal Police Office of Germany,
the identification error rate of 20.8\% was reduced to 2.1\% by rescaling.
The presented method is of striking simplicity
and can easily be integrated into existing automated fingerprint identification systems.

\end{abstract}

\paragraph{Keywords:} Fingerprint recognition, growth, matching, AFIS, shape analysis.

\section{Introduction and Prior Knowledge}

Identifying humans by their fingerprints has been a success story since its early origins in ancient China \cite{history}.
Nonetheless, the scientific foundation of fingerprint recognition still requires strengthening as demanded by the United States National Research Council in 2009 \cite{committeeforensicscience}. 
The situation is even worse when the fingerprints of juveniles are considered, which, in the same year,
lead the European Parliament to exempt children under 12 from having their fingerprints taken for visa purposes, consequently asking for a study to analyze the effects of growth on juveniles' fingerprints.
Annex 2, Article 2, 2nd paragraph in \cite{ecregulation} states:
\begin{quotation}\itshape
The first report shall also address the issue of the sufficient
reliability for identification and verification purposes of fingerprints of children under the age of 12 and, in particular, how
fingerprints evolve with age, based on the results of a study
carried out under the responsibility of the Commission.\\[6pt]
\end{quotation}

Indeed, most studies concerned with the effects of growth on fingerprints have focused on the stability of the line pattern's structure. 
Sir Francis Galton was among the first to demonstrate scientifically the permanence of the configuration of individual ridges and furrows in 1892~\cite{Galton1892}.
Subsequently, intensive pediatric research corroborated his findings, establishing that the pattern's development is finalized at an estimated gestational age of 24 weeks \cite{Babler1991}.

While postnatal growth of humans has been extensively studied \cite{methodshumangrowth},
especially its effects on stature and bone lengths \cite{humangrowth},
leading to growth charts widely used in pediatrics \cite{cdc},
up to now growth's effect on fingerprints appears to have escaped scientists' attention. 
There are only some genetically motivated, cross-sectional studies analyzing correlations between adult stature and dermatoglyphic characters, see e.g. \cite{dermatoglyphic}.
To the best of our knowledge there are no longitudinal studies of growth-related effects on children's and juveniles' fingerprints.

\begin{figure}[!tbp]
\noindent
\includegraphics[width=0.32\textwidth]{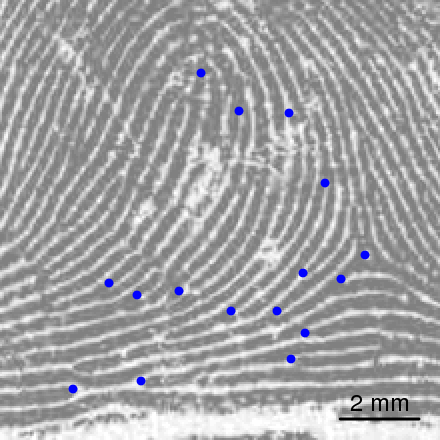}~
\includegraphics[width=0.32\textwidth]{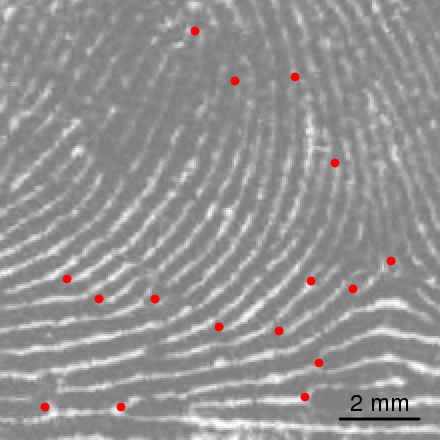}~
\includegraphics[width=0.32\textwidth]{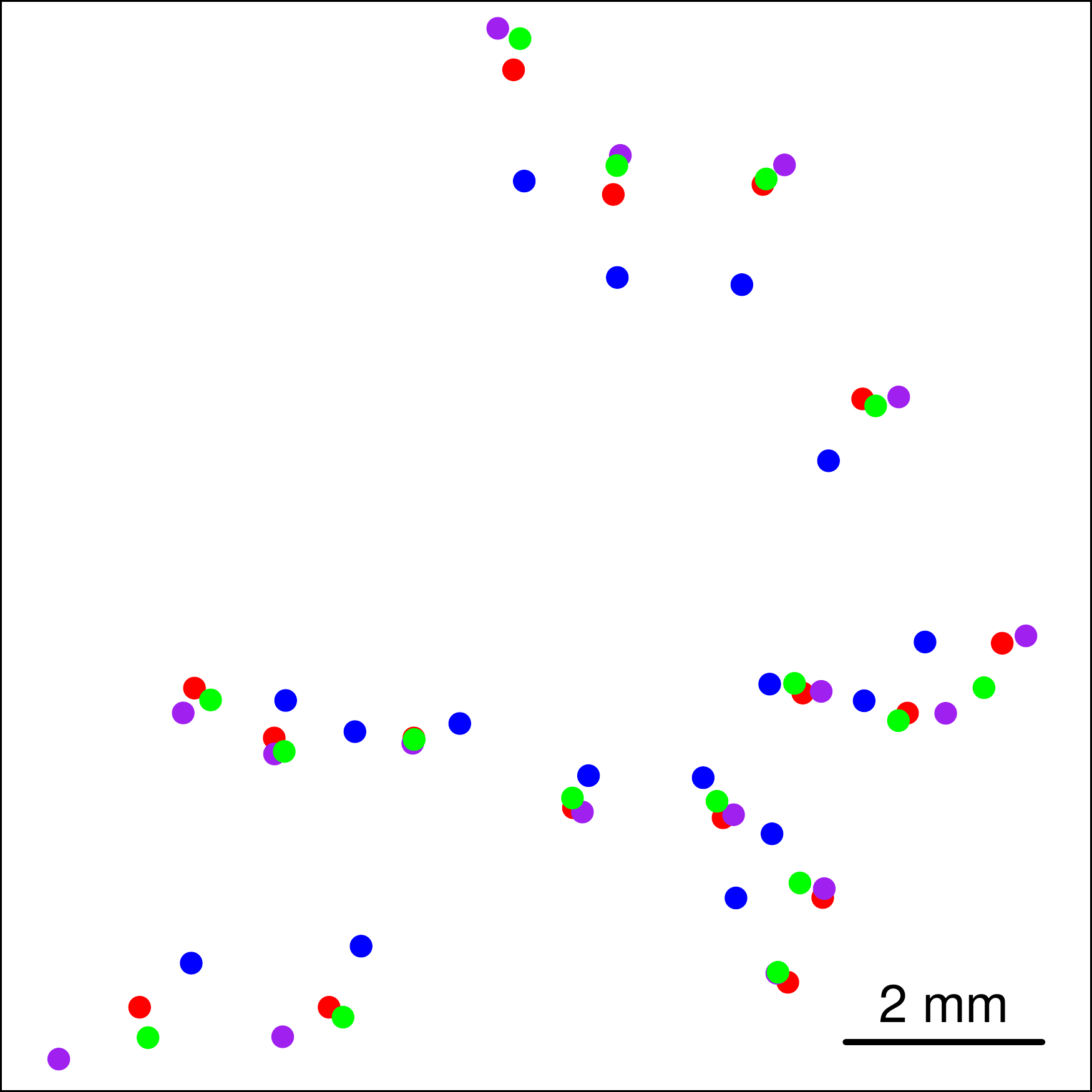}
\\[4pt]
\noindent
\includegraphics[width=0.32\textwidth]{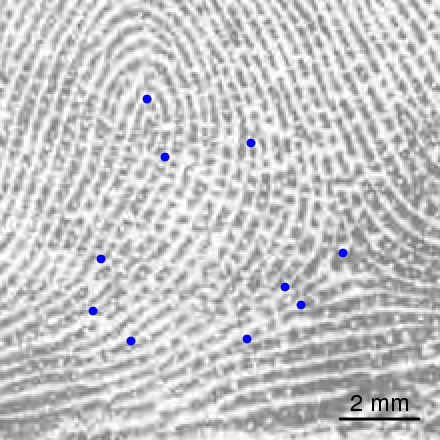}~
\includegraphics[width=0.32\textwidth]{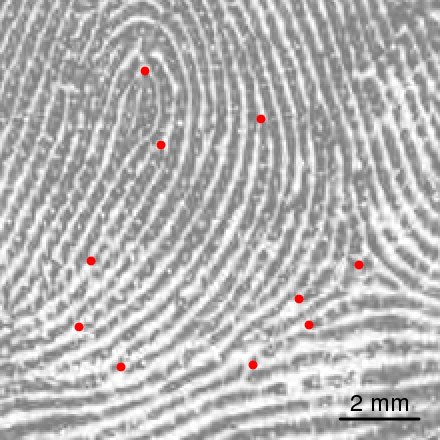}~
\includegraphics[width=0.32\textwidth]{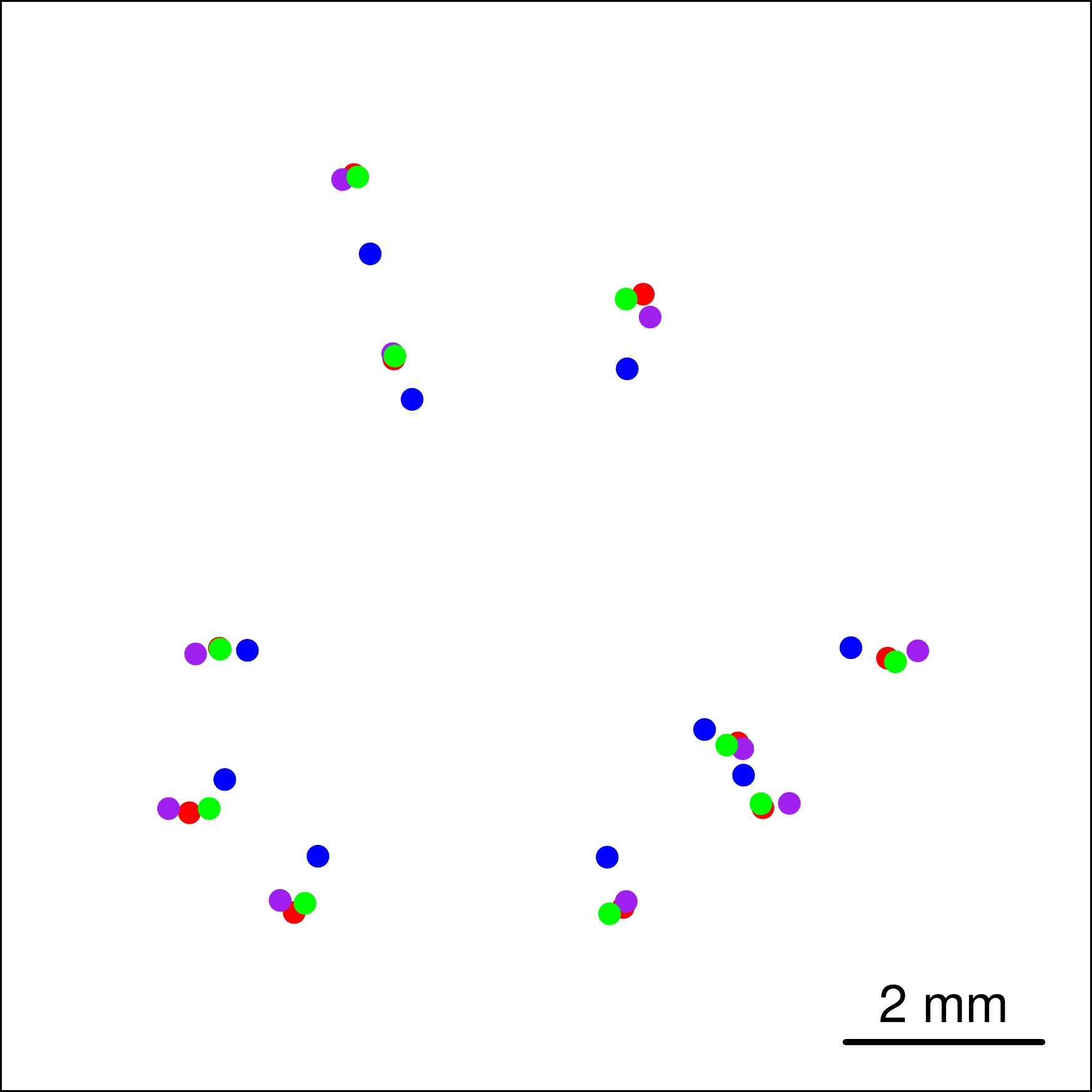}
\caption{\label{print}Imprints of right middle finger of persons 15 (top) and 18 (bottom) at first (left) and last (middle) check-out (CO) with points of interest (POI) marked by a human expert. Superimposed POI (right) at first CO without rescaling (blue), with rescaling (purple), each brought into optimal position w.r.t. the POI at last CO (red); control at last CO (green) shown for comparison (scale bars: 2 mm or 31.5 pixels).}
\end{figure}

As a finger grows, so does its skin expand,
effecting the relative position of the fingerprint's feature quantitatively. cf. Fig.~\ref{print}.
Hence, finger growth has also profound practical implications for law enforcement agencies:
if the person being checked out has been registered as a juvenile,
retrieving a matching fingerprint in their databases poses serious difficulties
to currently deployed automated fingerprint identification systems (AFIS).

The paper is organized as follows: in the next section, the available data source is described.
In Section \ref{secGrowth}, we introduce a measure for finger size and analyze finger growth
for anisotropic effects.
Based on these results we infer that fingers grow isotropically
and in Section \ref{secRescaling} we propose a method for adjusting prints before matching which can
easily be integrated into an existing AFIS.
The soundness and effectiveness of this approach is validated in Section \ref{secValidation} by conducting several tests.
The paper concludes with a summary and discussion, as well as an outline of topics for further research in Section~\ref{secDiscussion}.

\subsection{Data}

One explanation why growth effects on fingerprints have not been investigated yet may be the lack of longitudinal data sets of juveniles' fingerprints through adolescence.
For this study, the Federal Criminal Police Office of Germany (Bundeskriminalamt, BKA) provided fingerprints of reoffending juveniles that have been checked out in Criminal Records between 2 and 48 (median 4.5) times.
At their first check-out (CO), the subjects under study were 6--15 (median 12) years old, at the last CO 17--34 (median 25) years old; the 48 persons (35 male, 13 female) considered were born in 1972--1986.
The longitudinal data included information on birth date,
sex and date of CO such that age at CO could be determined.
At each CO a nail-to-nail rolled fingerprint was taken, as
well as an additional plain control.
The rolled imprint was used if not otherwise noted; the plain imprint
is called control wherever it was utilized.
The images used are scans of inked fingerprints on sheets of paper.
Image resolution was 500 DPI.


\section{Modeling Finger Growth}
\label{secGrowth}

Naturally, a child's finger increases in size, causing its skin to expand with it -- but are all parts of the fingerprint effected similarly, or is there possibly a dominant direction of growth, e.g. because fingers get rather elongated? Indeed, it is known that bones' ratio of length over width increases during growth \cite{elongate}. If the same happens to the fingerprint, distances along the bone would increase relative to distances orthogonal to it, requiring an elaborate correction step when comparing fingerprints obtained at different time points. Two effects therefore need to be disentangled: increase in size (or area) and changes in relative distances (or lengths) -- a common approach in auxology \cite{shape}.

\subsection{Shape Analysis}

To investigate the latter, the locations of points of interest (POI) were determined by a human expert in the fingerprint images 
(see Fig.~\ref{print}). POI are corresponding minutiae, these are bifurcations and endings of ridges, and singular points \cite{handbook}.
POI were marked in one finger for each person at all COs.
If at least one CO of the person was a so called simplified CO which consists only of the right index finger,
then this finger was chosen for that person. Otherwise, image quality at all COs was the decisive
criterion for choice of the finger. In order to maximize the number of POI identifiable at all COs, the finger with the largest overlapping area of acceptable quality was determined.
Fingers of the left and right hand were equally eligible,
since differences in proportions between bones of the left and right body side have been found insignificant \cite{penning}.

We defined the intra-finger measure of the pad \emph{size} to be
the square root of the sum of squared distances between 
the marked POI locations and their barycenter, i.e. what one may call the \textit{spread} of POI $S_{ijk}$:\\
\begin{equation}
\label{spread}
S_{ijk} = \sqrt{\sum_{m=1}^n d(M_{mijk}, \bar M_{\cdot{}ijk})^2}
\end{equation}
\\where $S_{ijk}$ is the pad size of person $i$'s marked finger, measured at CO $j$ from imprint $k$ (denoting the rolled imprint or the plain control),
and $d(M_{mijk}, \bar M_{\cdot{}ijk})$ is the Euclidean distance between the $m$-th minutia's coordinates $M_{mijk}$ and the minutiae's barycenter $\bar M_{\cdot{}ijk}$ in that print.
This measure allows to compare sizes of the same finger at different points in time.


In order to determine whether fingerprints grow isotropically, we employed tools from shape analysis \cite{shape,shapes,R}. For each marked finger for which more than two COs were available, we took the shape of POI in each imprint, i.e. we considered the POI only up to translation, rotation and rescaling. We then used full Generalized Procrustes Analysis (GPA) to represent each imprint as a point in a $2m-4$-dimensional Euclidean (tangent) space where $m$ denotes the number of POI marked for that finger. In that space, we computed the amount of variance explained by size (i.e. the spread of POI, used as a proxy for maturity) in a multivariate linear regression model containing no other regressor. By definition, the maximum amount any single regressor can explain in such a model is given by the variance explained by the first principal component (PC) in that tangent space, which we also computed. If growth deviates from isotropy, we expect both size and the first PC to explain much of the variance. We finally aggregated these per-person results by reporting their respective median values.

Similarly, partial GPA represents each imprint as a point in a high-dimensional Euclidean space but without factoring size out, i.e. only factoring translation and rotation out but distinguishing imprints if they differ in size. This allows to measure the effect of growth on both size and shape by following the same lines as for full GPA and the effect on shape alone.


For full GPA, we found the first PC to explain only a median 51\% of the total variation, size a mere 16\%. If, on the other hand, size was not factored out, i.e. for partial GPA, the first PC explained 64\% of the total variation, while size explained 58\%.
This suggests that the only strong effect growth has on fingerprints is a uniform increase in size,
otherwise a dominant direction of shape variation as well as a strong correlation with size would have been observed.
We conclude that during growth, fingerprints get isotropically rescaled, and all that remains to be determined is the scale factor.


\subsection{Rescaling}
\label{secRescaling}

With our forensic application in mind,
we need a method that allows to rescale a juvenile's fingerprint taken at age $x$ so that it matches his fingerprint taken at age $y$, $y - x$ years later. 
The scale factor is therefore only allowed to depend on the two ages $x$ and $y$ as well as on the juvenile's sex, but not on the person's unknown individual growth curve, i.e. we need a population-level growth chart for fingerprints.
As fingerprint growth charts are not available, we propose to use median curves of population-level growth charts for stature of boys and girls \cite{cdc} instead,
utilizing the strong correlation between stature and upper limb lengths, see e.g. \cite{humangrowth, himes}.


The scale factor $f$ depends on the two ages $x$ and $y$ at the times of CO, and the sex $s$ of the person:\\
\begin{equation}
\label{factor}
f(x,y,s) = \frac{m_{y,s}}{m_{x,s}},
\end{equation}
where $m_{x,s}$ denotes the median stature of the population with sex $s$ at age $x$.
The earlier print taken at age $x$ is then scaled up by the factor $f$ 
before matching with the later print which was acquired when the person was $y$ years old.



\begin{figure}[!tbp]
\begin{center}
\includegraphics[width=0.56\textwidth]{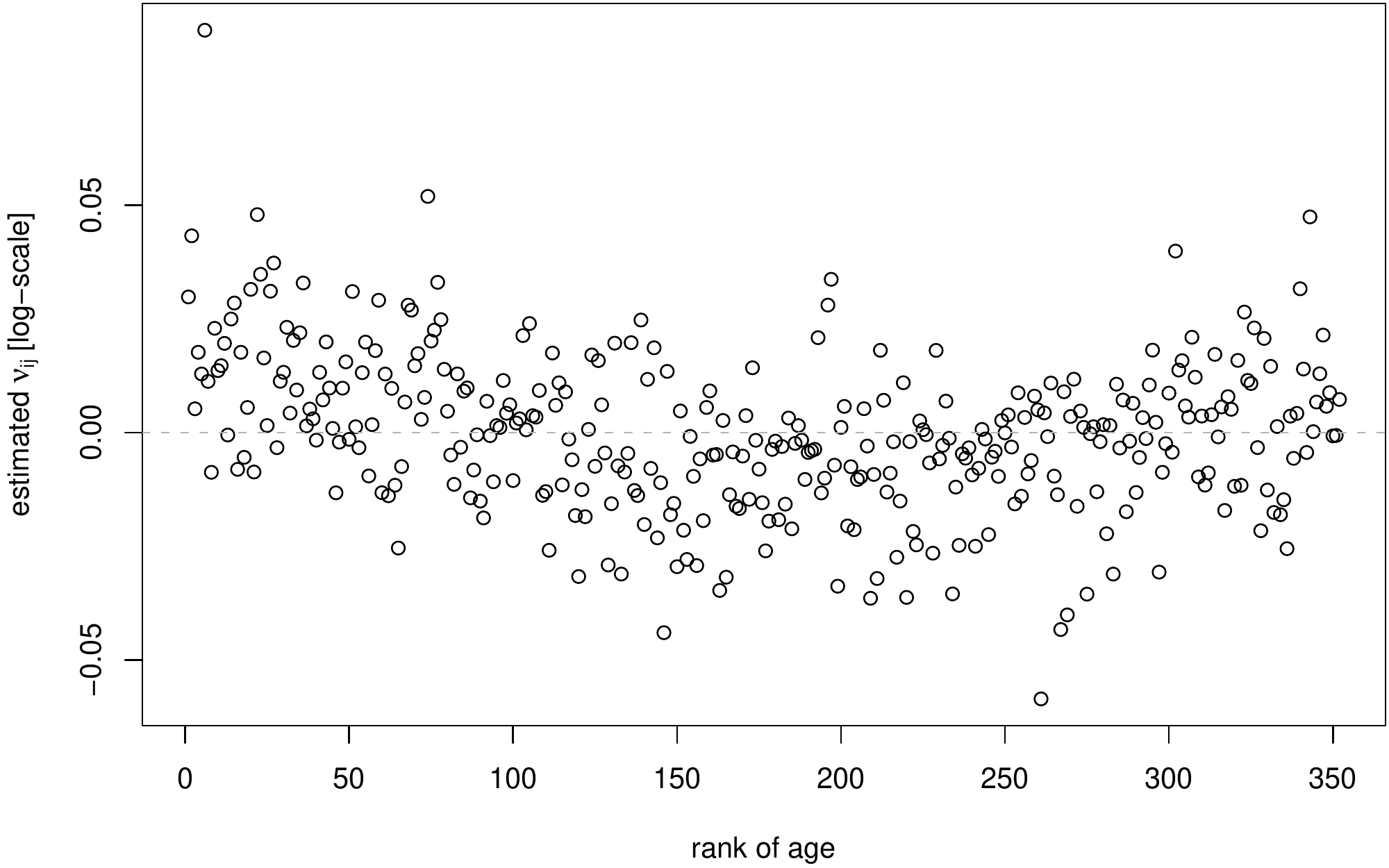}
\end{center}
\caption{\label{resid}Estimated random effects $\eta_{ij}$ showing persons' individual deviations from MGCS according to model (\ref{model}), displayed on the model's log-scale versus rank-transformed age of person $i$ at CO $j$.}
\end{figure}

\begin{figure}[!tbp]
\begin{center}
\includegraphics[width=0.42\textwidth]{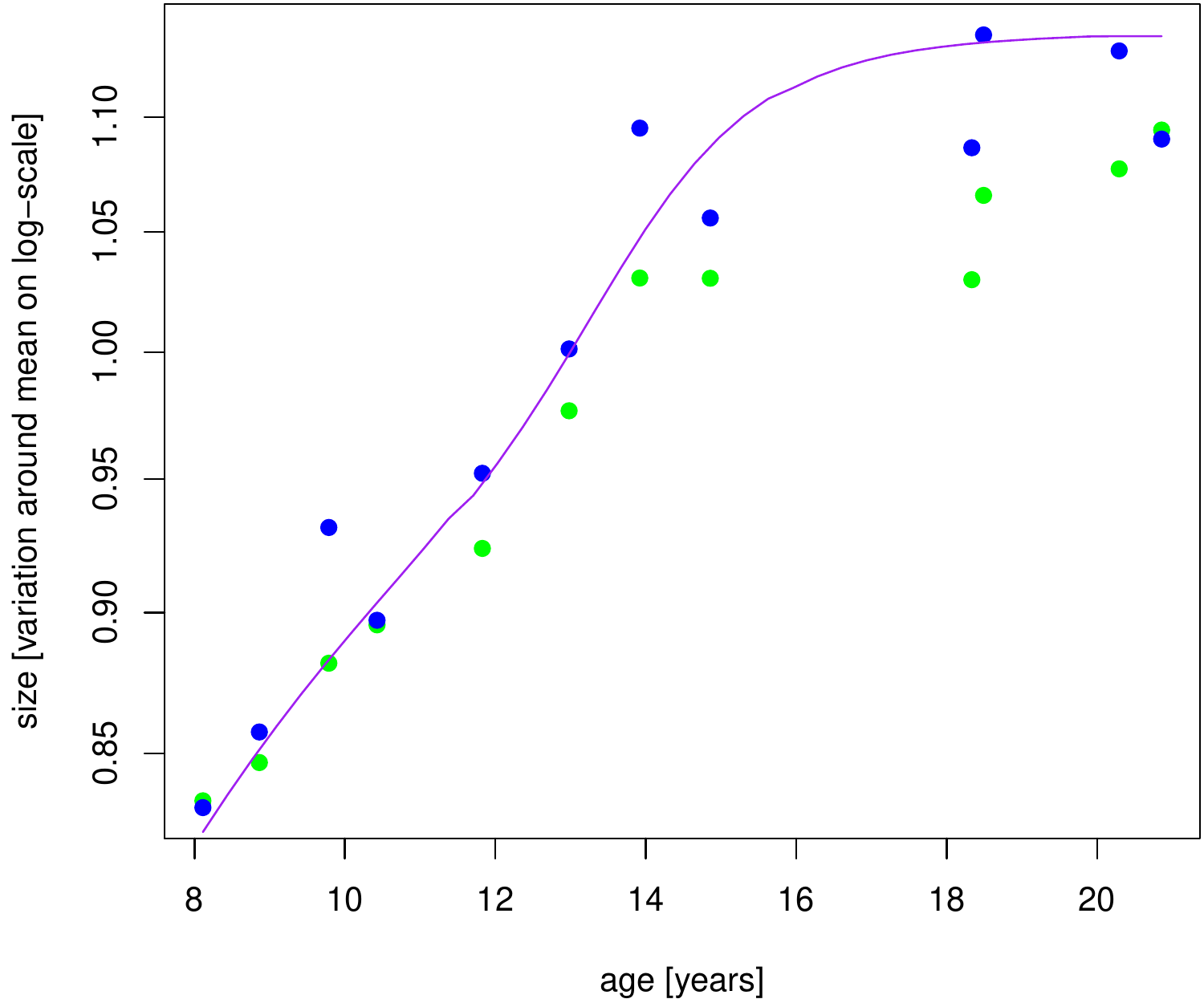}
\hspace*{6mm}
\includegraphics[width=0.42\textwidth]{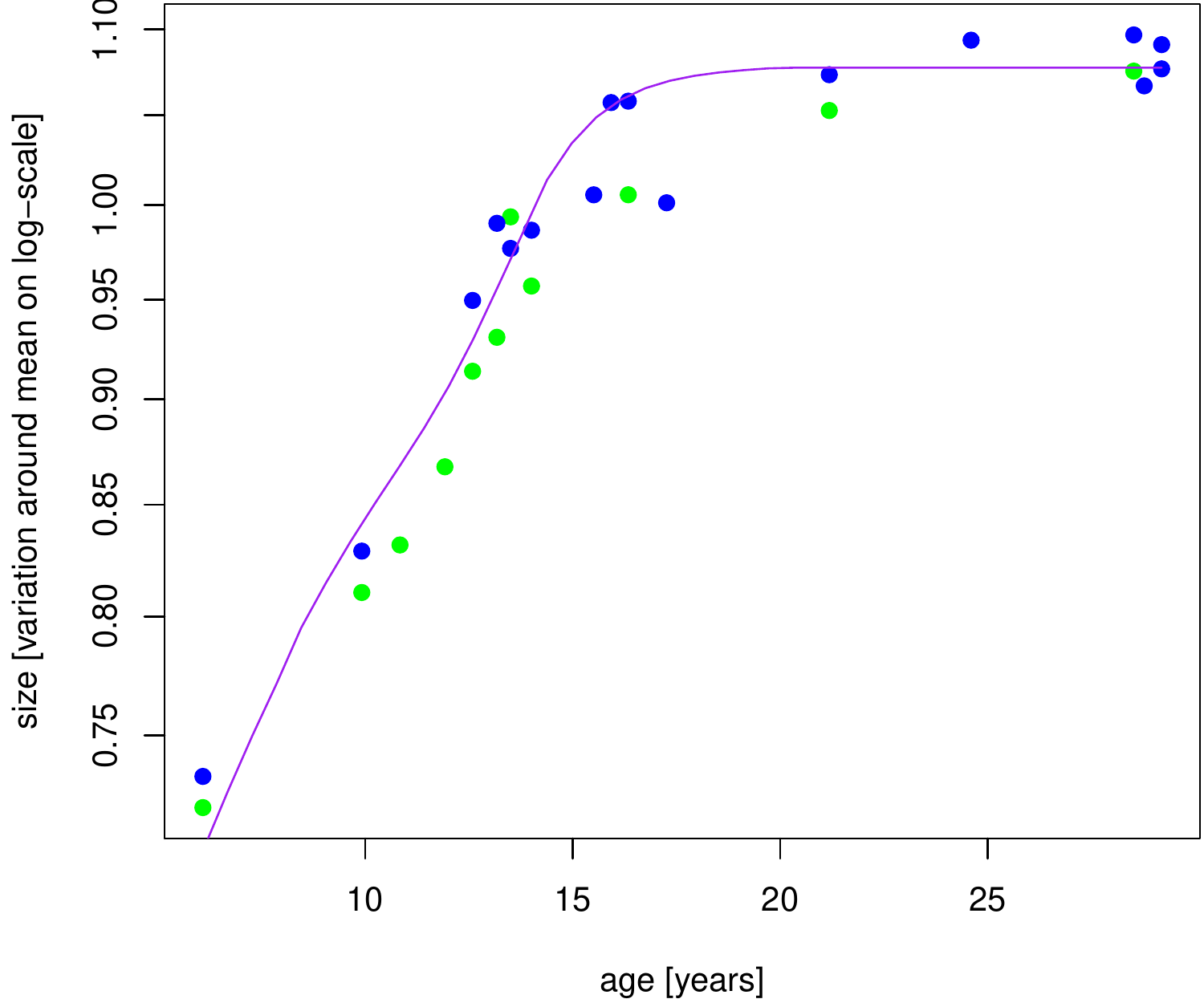}
\end{center}
\caption{\label{growth}Mean distance of POI at different COs (blue: original, green: control) of persons 17 (left) and 31 (right), as well as fitted growth according to growth chart (purple).}
\end{figure}

In order to investigate the adequacy of using median curves of stature growth for fingerprint sizes, 
we examined its fit to the spread of POI as an intra-finger measure of size.
%
More specifically, as growth has a multiplicative effect on size, we log-transformed the size $S_{ijk}$ of person $i$'s marked fingerprint $k$ at CO $j$ where $k$ denotes either the rolled imprint or the plain control, given in equation \eqref{spread}. The sizes were modeled proportional to the value $G_{ij} = m_{x_{ij},s_i}$ from the median curves of growth charts for stature (MGCS) for the corresponding age $x_{ij}$ and sex $s_i$ \cite{cdc}. To measure the deviances from MGCS, we considered the linear mixed effects model
\begin{equation}
\label{model}
\log S_{ijk} - \log G_{ij} = \mu_k + \eta_{ij} + \epsilon_{ijk},
\end{equation}
where $\mu_k$ is a fixed effect allowing for a systematic difference in the proportionality factors of rolled and plain imprints, while $\eta_{ij}$ and $\epsilon_{ijk}$ are independent random effects, Gaussian distributed with mean zero and standard deviations $\sigma_\eta$ and $\sigma_\epsilon$, resp. Here, $\eta_{ij}$ models person $i$'s systematic deviation from the MGCS, whereas $\epsilon_{ijk}$ models variations at the same time point, i.e. measurement noise, e.g. due to distortions when pressing the elastic skin on a flat surface \cite{deformation}. The model was fit by maximum likelihood \cite{nlme,R}, the estimated standard deviations were then transformed back into relative (or multiplicative) effects: 
we found the model misfit to have a standard deviation of about $\pm ( \exp(\hat\sigma_\eta) - 1 ) = \pm$2.26\%, similarly to the noise's standard deviation of about $\pm( \exp(\hat\sigma_\epsilon) - 1 ) = \pm$2.28\%,
the latter being due to non-linear distortions when the finger is pressed on a flat surface. We note that model misfit is inevitable as the prediction must be based on a population-level growth pattern which will fail to capture individual characteristics such as varying times of on-set of the pubertal growth spurt \cite{humangrowth}.
An exploratory residual analysis based on estimating the $\eta_{ij}$ reveals only small systematic deviations of the sampled populations' finger growth from MGCS, potentially due to comparably large fingers at younger ages, see Fig.~\ref{resid}.
Exemplary fits of the growth curve to the sizes of individual persons are shown in Fig.~\ref{growth}, obtained by demeaning on the log-scale and transforming back, i.e. by dividing the sizes by their geometric mean. These nicely visualize the strong correlation of fingerprint and stature growth.

Returning to our original problem, we now may propose a method for predicting a fingerprint at a later stage in growth based on an earlier one: compute a scale factor as the ratio of median stature for the corresponding ages and gender using equation \eqref{factor}, then uniformly scale the earlier print up by that factor. In practical applications, this may be easily achieved by accordingly reducing the images' DPI setting. An alternative to scaling the earlier print up is to scale the later print down; this 
leads to quantitatively similar results.
The quality and usefulness of this finger pad growth prediction was then examined in light of our forensic application using several verification and identification tests.

\section{Validation}
\label{secValidation}

\subsection{Minutiae Distances after Alignment}



For computing minutiae distances between two imprints of the same finger, the second fingerprint image was aligned to the first one using that rotation and translation which minimized the square root of mean of squared distances (SMSD) between the images' corresponding POI.
For each marked finger, we computed the SMSD of the imprints at first and last CO (rolled), reporting the SMSDs' median as a measure of typical mismatch. This was repeated with the imprint at first CO rescaled according to the MGCS; considering the ratio of the SMSD with and without rescaling for each person's marked finger gave the relative improvement per finger gained by rescaling. Comparing the plain control imprint to the rolled imprint, both at last CO, we obtained a measure of mismatch at the same time-point, i.e. excluding any growth effects. 

\begin{figure}
\begin{center}
\includegraphics[width=0.56\textwidth]{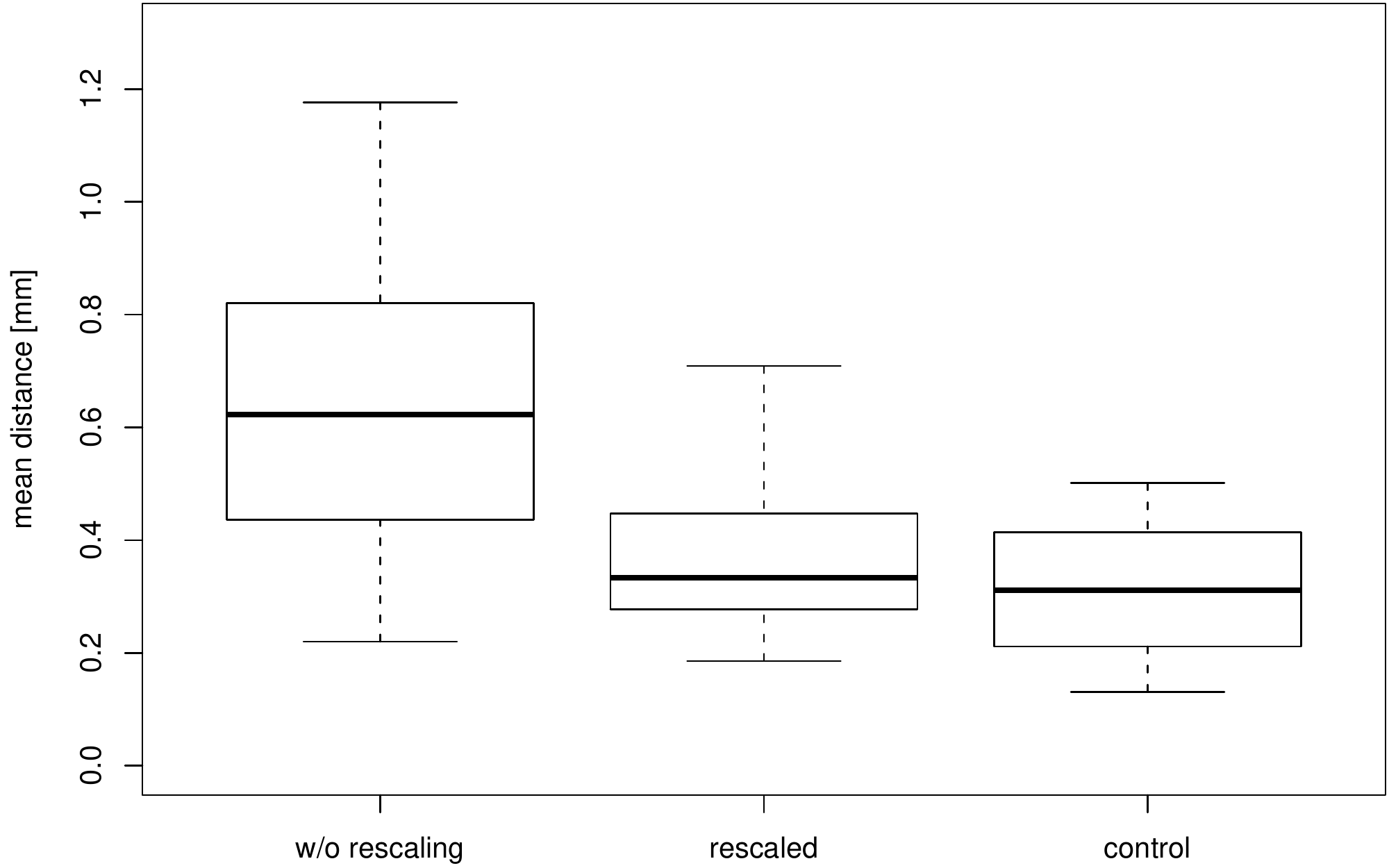}
\hspace*{6mm}
\includegraphics[width=0.35\textwidth]{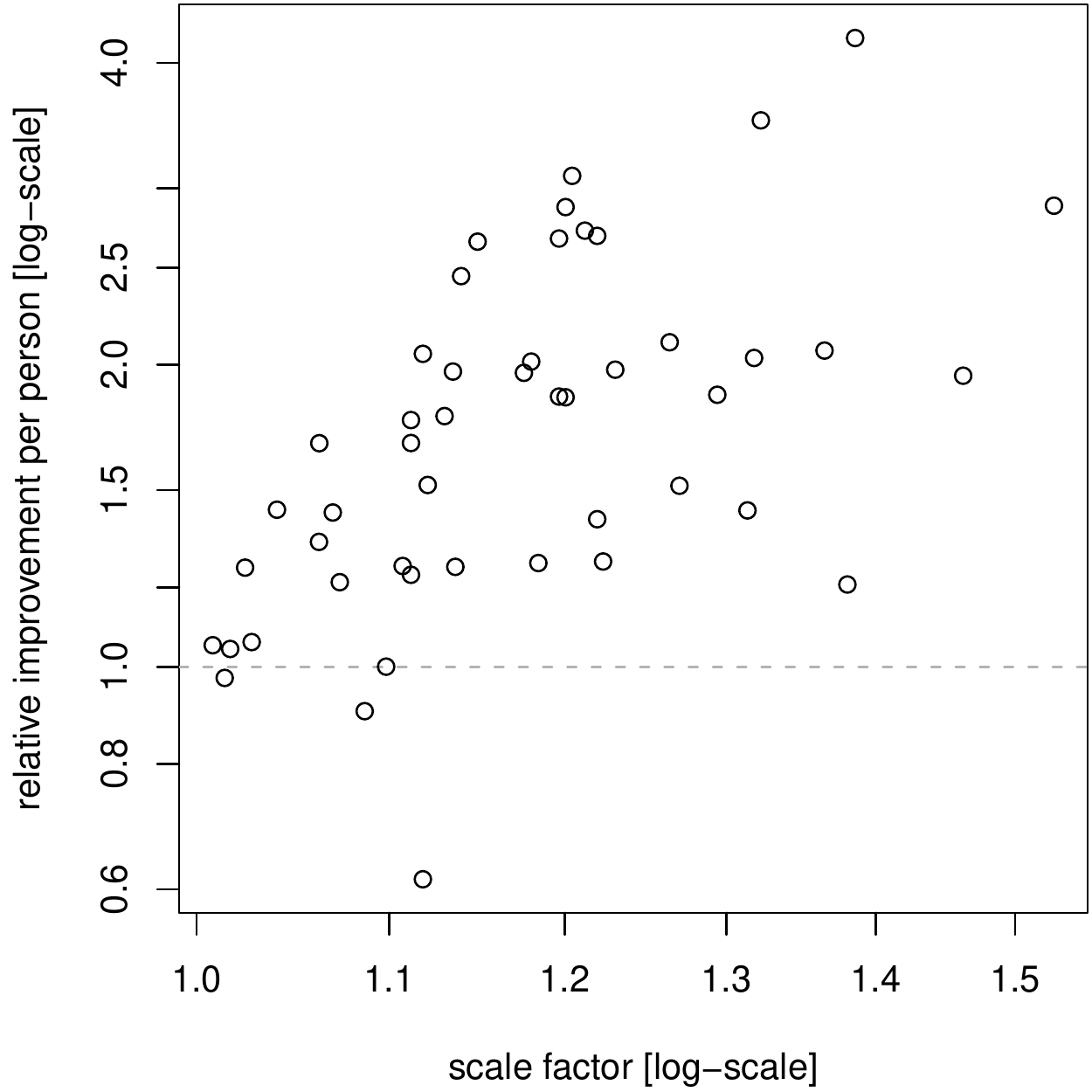}
\end{center}
\caption{\label{improve}Left: mean distance from POI at last CO after bringing POI at first CO without and with rescaling into optimal position, control at last CO shown for comparison, boxes show quartiles, whiskers 5\% and 95\% quantiles, resp.; cf. Fig.~\ref{print}. Right: relative reduction in this mean distance per person achieved by rescaling vs scale factor given by the growth chart.} 
\end{figure}

We found that rescaling can reduce the median distance from 0.78~mm when growth is not taken into account to 0.42~mm with the suggested method; for comparison, the median distance between the two prints at the last CO was 0.39~mm, which is indicative of the achievable accuracy;
see Fig.~\ref{improve} (left).
Clearly, the improvement per finger depends on the years that have passed, more precisely on the amount  of growth as given by the scale factor; indeed there exists a strong correlation between the two, see Fig.~\ref{improve} (right).

\subsection{Verification Tests}

\label{matcher}

Since fingerprint matching algorithms are typically based on the distance of matched POIs, we expect similar improvements in matchers' performance, too.
To ensure that the results do not depend on the particular matching algorithm used, four different matchers were applied: 
a free algorithm from the NIST, two variants of a commercial software and the BKA's in-house matcher.
The matcher referred to as ``Bozorth3'' is based on the NIST biometric image software package \cite{nist}, applying algorithm MINDTCT for minutiae extraction and BOZORTH3 for template matching.
Two other matchers, named ``Verifinger Grayscale'' and ``Verifinger Minutiae'' have been created using the Neurotechnology VeriFinger 5.0 SDK.
The first variant directly matches two grayscale images, the second extracts minutiae, stores them into a template
and matches the templates.
For the two minutiae based algorithms ``Bozorth3'' and ``Verifinger Minutiae'',
the rescaling was performed by adjusting the coordinates in the template files, whereas the fingerprint image was rescaled for ``Verifinger Grayscale'' and the BKA's in-house matcher.

To assess the practical relevance of the rescaling, we determined the false rejection and false acceptance rates (FRR and FAR, respectively) in a verification test using different matching algorithms; in this scenario, the matcher must decide whether a query fingerprint belongs to the same person who registered previously to the database. 
As FAR and FRR are functionally related, usually via a threshold parameter for the scores, we determined equal error rates (EER) \cite{EERdefinition}. 

The verification tests were conducted according to the following protocol:
all available fingerprints of the first and last COs were used in the test.
For 2 of the 48 persons, only the right index finger could be considered due to a simplified CO,
whereas for the other persons all ten fingers were available, resulting in 462 different fingers.
For each fingerprint of the last CO, one genuine recognition attempt was performed by matching it against the same finger's imprint at the first CO,
and 461 impostor recognition attempts were conducted by matching it against all other fingerprints of the last CO.
This protocol corresponds to a scenario with a database containing a vast majority of adult fingerprints and only few fingerprints of children and juveniles,
as it is the case at the BKA.
In order to measure the effects of rescaling, the genuine fingerprint at first CO was once left unscaled and once rescaled according to MGCS.
Since this protocol includes impostor recognition attempts with prints of different fingers from the same person,
matching accuracy is expected to be slightly lower in comparison to matching genetically independent samples,
analogous to results of verification tests with fingerprints of identical twins \cite{twins}.
Scores were normalized using median and MAD (see p. 327 in \cite{handbook}) before fusion
and all 48 persons were weighted equally by averaging error rates within persons first.
Thereby we corrected for the fact that prints of ten fingers could be used for 46 persons,
while for two persons only one finger was available; however, we note that this does not significantly affect the error rates.

\begin{table}[!tbp]
\begin{center}
\begin{tabular}{l|rr}
\emph{Equal error rates} & without rescaling & with rescaling \\[3pt]
\hline&&\\[-9pt]
NIST Bozorth3 & 13.3 \% & 6.4 \% \\
Verifinger Grayscale & 14.1 \% & 5.3 \% \\
Verfinger Minutiae & 10.8 \% & 5.4 \% \\
Sum Rule &  6.8 \% & 3.1 \% \\
3 Factors &  6.8 \% & 1.8 \% 
\end{tabular}
\end{center}
\caption{\label{eer} Equal error rates using all fingers.}
\end{table}

Without rescaling, we obtained EERs in the range from 
10.8 to 14.1\%. 
EERs are cut in half by magnifying the first finger 
(5.3--6.4\%). 
Score fusion using the simple sum rule \cite{scorefusion} leads to a further decrease of the EERs:
6.8\% without and 3.1\% with rescaling (see Table \ref{eer}).

The use of growth charts' median curves works well for the vast majority of cases examined.
However, as individual growth may deviate considerably from average growth, trying to match with multiple, differently rescaled versions of a finger might result in higher matching scores in such a case. 
Indeed, additionally considering $\pm 5$\% of the scale factor predicted by the median curve and using the maximum score of these variations leads to an EER of 1.6\% for the sum rule; we called this method ``3 Factors''.
Visual inspection found the remaining matching errors to be caused by very low quality of the fingerprints.
Using multiple rescaled fingerprints thus overcomes the difficulties of predicting individual growth from one single data set.

\subsection{Identification Tests}

\begin{table}
\hspace*{-2mm}\parbox{0mm}{
\begin{tabular}{l|rr|rr}
\raisebox{-6pt}[0pt][0pt]{\emph{Proportion of}} & \multicolumn{2}{c}{Top Rank} \vline& \multicolumn{2}{c}{One of Top 3 Ranks} \\
& without rescaling & with rescaling & without rescaling & with rescaling \\[3pt]
\hline&&&&\\[-9pt]
NIST Bozorth3  &  59.1  \% &  82.5  \%   &  67.5  \% &  85.9  \% \\
Verifinger Grayscale  &  64.3  \% &  90.5  \%   &  69.5  \% &  92.4  \% \\
Verfinger Minutiae  &  69.7  \% &  90.7  \%   &  75.3  \% &  92.0  \% \\
Sum Rule  &  81.3  \% &  94.6  \%   &  85.4  \% &  95.4  \% \\
3 Factors  &  81.3  \% &  96.9  \%   &  85.4  \% &  97.7  \% \\
BKA  &  79.2  \% &  97.9  \%   &  79.2  \% &  97.9  \% \end{tabular}
}
\caption{\label{rank} Proportion of genuines giving the highest score or one of the three highest scores, resp.}
\end{table}

From a forensics point of view, verification is less of an issue; 
what matters more is identification \cite{jainnature}:
whether a search in the database given some query fingerprint will return the same person's juvenile fingerprint as part of the top ranked search results. We therefore also determined how often a fingerprint at last CO gave the highest score when comparing it to all fingers at first CO using the matchers described above.
Furthermore, the same identification test was conducted at the BKA but searching in a database containing 3.25 million right index fingers, including the rescaled or unscaled right index fingerprint at first CO of all 48 subjects, and querying the database with each subject's right index finger obtained at last CO.
The unscaled fingerprint showed up as one of the top three results in 38 out of 48 cases, meaning that 10 persons would not have been identified. Rescaling the fingerprint at first CO resulted in the highest score in 47 cases,
see Table~\ref{rank}.
Only one person could not be identified with either scaling;
visual inspection of the right index finger's imprint at first CO showed the image disturbed by smudges 
while its imprint at last CO had been exceedingly affected by nonlinear distortions.
If the left index finger's imprints had been used instead,
both scalings would have led to a top rank matching result.


\section{Summary and Discussion}
\label{secDiscussion}

Summarizing these results, the analysis of growth effects on fingerprints clearly showed growth to result chiefly in an isotropic rescaling. This insight yielded a simple, practical solution for dealing with juvenile fingerprints in law enforcement agencies' databases: predict the growth using boys' and girls' growth charts by equation \eqref{factor} and rescale the database's fingerprints accordingly, potentially employing multiple templates which account for deviations from the median growth. This can easily be integrated into existing AFIS; regular, e.g. quarterly, database updates can adjust 
juvenile fingerprint templates accordingly, keeping them up-to-date. Performing the updates at times of low workload or offline renders this procedure technically feasible even for databases with millions of entries.

The effectiveness of the proposed method for growth prediction was verified in three tests: first we demonstrated that the distances of corresponding POIs were halved by rescaling; secondly, EERs on a test set of 462 fingers were also halved; and thirdly, this result result was confirmed on the BKA's database comprising 3.25 million right index fingers on which 9 failures to retrieve a juvenile fingerprint out of 48 such identification attempts could be avoided by rescaling.

While the exact number of failures to identify persons that currently occur because the impact of growth on fingerprints is ignored remains unknown, we know of cases in which the records of the first and last CO could not be matched despite all ten of the person's fingerprints being used; the identifications of these persons was only possible through intermediate COs and chain inference. With the proposed method, we hope to enable law enforcement agencies to reduce the dark figure of missed identifications.

We note that during the first statistical analysis of model misfit, we detected two clear outliers. Investigating these more closely, we discovered that in one case the date of CO was indeed off by 10 years due to a typing error. In the other case, the true year of birth is unknown since different birth dates were given at different COs, probably to avoid punishment by understating the age; the fingerprint's growth data appear to underpin this supposition. After correcting these two entries for the final analysis, both outliers disappeared, and the data were consistent with the growth model.

Although the acquired evidence speaks strongly in favor of using growth charts of stature to model fingerprint's growth, reliable growth charts for adolescents' fingers nonetheless remain desirable from a scientific point of view. When using multiple scale factors, e.g. method ``3 Factors'', however, their practical relevance may be limited.
Moreover, only 25 subjects were under 12 years old at first CO,
thus the question of growth effects on fingerprints of children that age, as asked by the European Parliament, has not been fully answered yet.
Noting that stature grows faster than the hand in early childhood, the corresponding scale factors might also differ; however, we conjecture that fingers also grow uniformly then.
Also, we reckon that matching could further be improved upon if the true body heights were available at the COs, by using these to obtain a refined scale factor.
These topics certainly deserve further research.

\end{document}